\documentclass[journal]{IEEEtran}

\usepackage{amsmath,amssymb,amsfonts}
\usepackage{graphicx}
\usepackage{booktabs}
\usepackage{array}
\usepackage{cite}
\usepackage{url}
\usepackage{balance}
\usepackage{tikz}
\usetikzlibrary{arrows.meta,positioning,calc,fit,backgrounds,shapes.geometric}

\graphicspath{{figures/}}

\tikzset{
  joint/.style={circle,draw,fill=white,minimum size=4pt,inner sep=0pt},
  link/.style={line width=2.2pt,line cap=round,gray!75},
  blk/.style={draw,rounded corners=2pt,align=center,font=\footnotesize,
              minimum height=7mm,inner sep=3pt,fill=blue!4},
  est/.style={draw,rounded corners=2pt,align=center,font=\footnotesize,
              minimum height=7mm,inner sep=3pt,fill=orange!8},
  sig/.style={-{Latex[length=2mm]},line width=0.6pt},
  sumc/.style={circle,draw,minimum size=4.5mm,inner sep=0pt,font=\small}
}

\newcommand{\Ieff}{I_f^{\text{eff}}}
\newcommand{\beff}{b_f^{\text{eff}}}
\newcommand{\Fmpc}{F_{\text{mpc}}}
\newcommand{\Fff}{F_{\text{ff}}}
\newcommand{\Fmpck}[1]{F_{\text{mpc},#1}}
\newcommand{\Fffk}[1]{F_{\text{ff},#1}}
\newcommand{\dhat}{\hat{d}}

\begin{document}

\title{Interaction Dynamics for Dexterous Manipulation}

\author{Yongyan~Cao%
\thanks{Y. Cao is with Voryx Robotic LLC, San Jose, CA 95136, USA
(e-mail: yongyancao@gmail.com).}%
\thanks{Preprint, June 2026. This work has been submitted for possible publication; copyright may be transferred without notice.}}

\markboth{Preprint}%
{Cao: Interaction Dynamics for Dexterous Manipulation}

\maketitle

\begin{abstract}
Physical AI systems must continuously predict and regulate \emph{interaction dynamics}---the closed-loop relation among commanded
motion, contact force, and actuation limits---under real-time safety constraints, yet the interaction layer beneath learned policies
is still left to fixed-gain controllers with no explicit physical model. We treat dexterous manipulation as a \emph{model-based}
interaction-dynamics problem: rather than track motion and react to contact, the controller predicts and regulates the interaction
itself. A sustained contact torque $\tau_{\text{ext}}$ through a joint
stiffness $K_d$ produces the structural bias $e_\infty=\tau_{\text{ext}}/K_d$, so any fixed-gain controller trades accuracy against
contact safety. We make these interaction dynamics explicit and
actuator-agnostic through a constant-$A_d$ double-integrator backbone, instantiating the offset-free architecture established for
physical human--robot interaction (pHRI) and preserving its modeling assumptions on the reduced residual dynamics. An algebraic feedforward reduces any tendon transmission to a constant-coefficient double integrator, so the QP cost inverse is
precomputed offline and a 10-step receding-horizon QP runs at 500\,Hz under contact-force (ISO/TS 15066), actuation, and jerk 
constraints. An encoder-only augmented-Kalman disturbance state drives steady-state error to zero under constant contact loads 
in the nominal detectable case. In simulation, a hydraulically actuated finger---the worked example, adding pressure and cavitation 
constraints---attains 0.6\,mrad RMS, 0.1\,mrad steady-state, and 7.3\,mrad peak deflection under 1.5\,Nm contact: 153$\times$, 
1500$\times$, and 21$\times$ better than classical impedance. The realized first-move stiffness (18$\to$323\,Nm/rad with update rate) 
is independently verified, and the architecture scales to a 16-DOF LEAP Hand MuJoCo model, recovering from 2.5\,N grasp disturbances
within 0.7\,s.
\end{abstract}

\begin{IEEEkeywords}
Dexterous manipulation, impedance control, model predictive control, Kalman filter, disturbance rejection, tendon-driven hands.
\end{IEEEkeywords}

\section{Introduction}

\IEEEPARstart{P}{hysical} AI---embodied systems that act in the physical world rather than purely in digital environments---requires
continuous reasoning over interaction dynamics, contact, and uncertainty under real-time safety constraints. Such systems increasingly
lean on learned policies, while the underlying interaction dynamics are handled by low-level controllers that carry no explicit
physical model, leaving the contact layer the least certified and least data-efficient part of the stack. We argue that physical
intelligence also needs a \emph{model-based} interaction layer---one that predicts interaction dynamics, enforces physical
constraints, and remains stable during contact---and this paper develops one such layer for dexterous manipulation. Because it reduces contact to a constant analytic model, this layer doubles as an interaction-dynamics prior for model-based physical AI, on which a learned component need only capture the residual uncertainty of contact and object.

Dexterous manipulation is, at its core, a problem of \emph{interaction dynamics}: the closed-loop relation among
commanded finger motion, the contact force exchanged with a grasped object, the actuation and pressure limits of the transmission,
and the safety filters that bound them. Robotic hands route actuation to
the phalanges through tendon transmissions---hydraulic, cable, pneumatic, twisted-string, or series-elastic---each of which introduces
reflected inertia, transmission damping, and platform-specific nonlinearities that complicate high-bandwidth finger control, while safe
operation with grasped objects and human co-workers demands predictable, bounded contact forces (the ISO/TS 15066 limit of 140\,N at
robot--human contact points \cite{iso15066}). Throughout, a hydraulically actuated finger serves as the worked example: hydraulics offers the highest force density yet is the most demanding transmission to control---coupled 
pressure dynamics, configuration-dependent inertia, compressibility resonance \cite{merritt1967}. The approach is actuator-agnostic;
Section~\ref{sec:tendon} gives the substitutions for the other architectures.

Classical impedance control \cite{hogan1985} resolves the accuracy--safety trade-off by shaping the mechanical port behavior at the joint, 
but stiffness and disturbance rejection cannot be optimized simultaneously: for $K_d = 10$\,Nm/rad and $\tau_{\text{ext}} = 1.5$\,Nm the 
steady-state error is $e_\infty = \tau_{\text{ext}}/K_d = 150$\,mrad, independent of loop bandwidth, while integral action (PI impedance) 
introduces windup across sequential contact events \cite{cao2004antiwindup}, \cite{cao2002antiwindup}. Admittance control \cite{siciliano1999}, 
\cite{haddadin2009} instead yields toward contact---correct for rehabilitation and pHRI, but wrong for autonomous grasping, where contact is 
a disturbance to be rejected. MPC can resolve both objectives---predictive constraint enforcement plus offset-free disturbance rejection
through an augmented estimator state \cite{rawlings2017}---but existing impedance-MPC formulations forfeit either the update rate or
per-step offset-free rejection under persistent contact (Section~\ref{sec:related}); none reaches the sub-15\,mrad precision window
required for dexterous finger positioning under contact.

\begin{figure*}[t]
\centering
\begin{minipage}[c]{0.36\textwidth}
\centering
\includegraphics[width=0.98\linewidth]{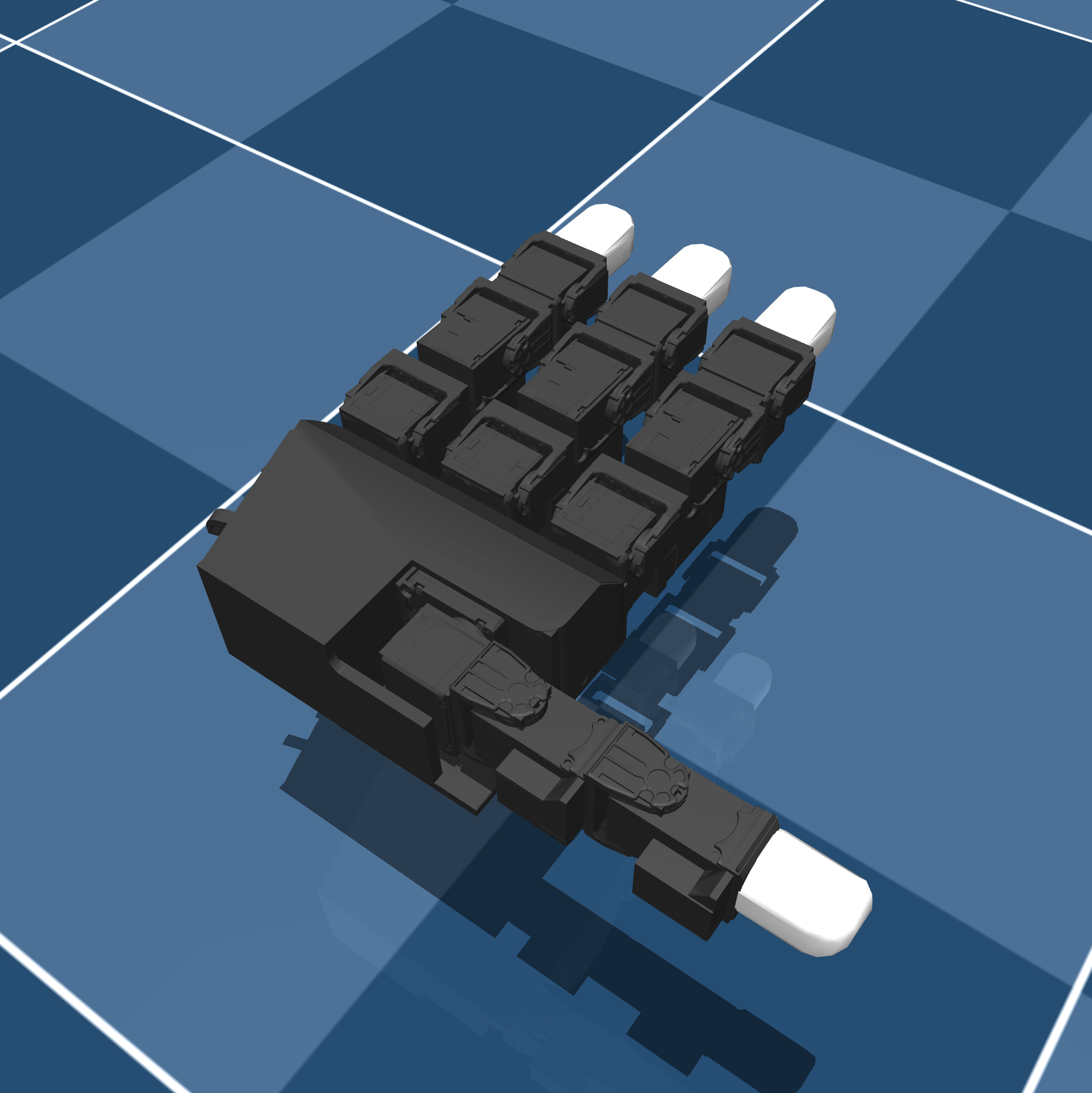}\\[-1mm]
{\scriptsize (a) MuJoCo LEAP-class hand model used in the full-hand study.}
\end{minipage}\hfill
\begin{minipage}[c]{0.60\textwidth}
\centering
\begin{tikzpicture}[scale=0.92,every node/.style={font=\footnotesize}]
  \fill[gray!12,draw=gray!55,rounded corners=2pt] (-1.7,-0.7) rectangle (1.7,0.3);
  \node at (0.2,-0.2) {\scriptsize palm};
  \foreach \bx/\lab in {-1.3/I,-0.3/M,0.7/R}{
    \coordinate (\lab0) at (\bx,0.30);   
    \coordinate (\lab1) at (\bx,1.10);   
    \coordinate (\lab2) at (\bx,1.80);   
    \coordinate (\lab3) at (\bx,2.35);   
    \draw[link] (\lab0)--(\lab1)--(\lab2)--(\lab3);
    \node[joint] at (\lab0){}; \node[joint] at (\lab1){}; \node[joint] at (\lab2){};
  }
  \coordinate (T0) at (-1.7,-0.35);
  \coordinate (T1) at (-2.45,0.1);
  \coordinate (T2) at (-2.95,0.7);
  \coordinate (T3) at (-3.25,1.2);
  \draw[link] (T0)--(T1)--(T2)--(T3);
  \node[joint] at (T0){}; \node[joint] at (T1){}; \node[joint] at (T2){};
  \node[below left=0pt and -2pt of T0,font=\scriptsize] {thumb (4 DOF)};
  \node[right=2pt of R0,font=\scriptsize] {MCP\,($\theta_f$)};
  \node[right=2pt of R1,font=\scriptsize] {PIP};
  \node[right=2pt of R2,font=\scriptsize] {DIP};
  \draw[-{Latex[length=1.4mm]},gray!70] (0.7,0.30) ++(0,-0.30) arc (270:320:0.30);
  \node[below right=-1pt and 1pt of R0,font=\scriptsize,gray!60] {abd/rot};
  \draw[dashed,red!60] (-1.55,0.18) rectangle (-1.05,2.4);
  \node[red!70,font=\scriptsize,align=center] at (-1.3,2.75)
        {hydraulic\\ worked example};
\end{tikzpicture}
\vspace{-1mm}

{\scriptsize (b) Kinematic abstraction: four 4-DOF digits. Each finger has an abduction/rotation DOF and three flexion joints 
(MCP, PIP, DIP); $\theta_f$ denotes a controlled joint angle.}
\end{minipage}
\caption{Dexterous hand studied in this paper. The physical hardware is not used in this manuscript; the full-hand results are 
obtained with the MuJoCo LEAP-class model in (a), while (b) defines the joint structure and the single-finger worked example 
whose transmission is detailed in Fig.~\ref{fig:hydraulic}.}
\label{fig:hand}
\end{figure*}

\begin{figure}[t]
\centering
\resizebox{\columnwidth}{!}{%
\begin{tikzpicture}[scale=1.0,every node/.style={font=\footnotesize}]
  \node[draw,fill=gray!15,minimum width=8mm,minimum height=8mm] (mot) at (0,0) {M};
  \node[below=0pt of mot,font=\scriptsize] {motor};
  \draw (1.0,-0.45) rectangle (2.2,0.45);
  \fill[blue!10] (1.0,-0.45) rectangle (1.9,0.45);
  \draw[fill=gray!40] (1.55,-0.45) rectangle (1.7,0.45);
  \draw[sig] (mot.east)--(1.55,0);
  \node[below,font=\scriptsize] at (1.6,-0.5) {$A_1,P_1$};
  \node[above,font=\scriptsize] at (0.78,0.0) {$F$};
  \draw[-{Latex[length=1.6mm]}] (1.62,0.62)--(2.05,0.62);
  \node[above,font=\scriptsize] at (1.83,0.66) {$x_1$};
  \draw[line width=3pt,blue!18] (1.9,-0.25)--(3.3,-0.25);
  \draw (1.9,0.0)--(3.3,0.0);
  \draw (3.3,-0.45) rectangle (4.6,0.45);
  \fill[blue!10] (3.3,-0.45) rectangle (4.1,0.45);
  \draw[fill=gray!40] (3.75,-0.45) rectangle (3.9,0.45);
  \node[below,font=\scriptsize] at (3.95,-0.5) {$A_2,P_2$};
  \draw[-{Latex[length=1.6mm]}] (3.95,0.62)--(4.38,0.62);
  \node[above,font=\scriptsize] at (4.16,0.66) {$x_2$};
  \draw[link] (4.55,0)--(5.35,0);
  \node[joint] (jt) at (5.35,0){};
  \node[below=2pt of jt,font=\scriptsize] {$J_f$};
  \draw[link] (jt)--(6.45,0.85);
  \node[joint] at (6.45,0.85){};
  \draw[link] (6.45,0.85)--(7.15,1.7);
  \draw[-{Latex[length=1.4mm]}] (6.05,0) arc (0:38:0.7);
  \node[font=\scriptsize] at (6.32,0.32) {$\theta_f$};
  \draw[-{Latex[length=1.8mm]},red!75,line width=0.9pt] (7.6,2.15)--(7.15,1.7);
  \node[red!75,font=\scriptsize,align=center] at (8.05,2.3) {$F_{\text{ext}}$\\($\tau_{\text{ext}}$)};
\end{tikzpicture}}
\caption{Hydraulic tendon transmission of one finger (the worked example),
mapping the variables of this section onto the physical parts: motor drive
force $F$, master/slave piston areas $A_1,A_2$ and pressures $P_1,P_2$, piston
displacements $x_1,x_2$, mechanism Jacobian $J_f$, joint angle $\theta_f$, and
the external contact force/torque $F_{\text{ext}}/\tau_{\text{ext}}$. The other
tendon architectures of Table~\ref{tab:tendon} replace this stage while leaving
the controller of Fig.~\ref{fig:block} unchanged.}
\label{fig:hydraulic}
\end{figure}
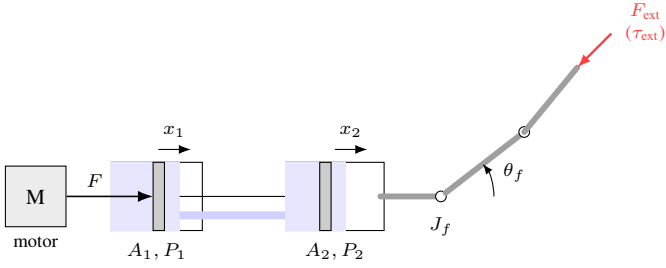

\emph{Problem statement.} Consider one digit of the dexterous hand of
Fig.~\ref{fig:hand}, whose joint angle $\theta_f$ is driven through a tendon
transmission (Fig.~\ref{fig:hydraulic}) from a single actuator and measured by
a joint encoder; the variables are collected in Table~\ref{tab:nomen}. An
\emph{a priori} unknown, time-varying contact torque $\tau_{\text{ext}}$ acts
at the fingertip during grasping. The objective is a single controller that,
using \textbf{joint encoders alone} (no force/torque sensor), simultaneously:
(i)~tracks a reference trajectory $\theta_d(t)$ to within the dexterous-grasping
precision window of $15$\,mrad; (ii)~maintains that precision through the onset
\emph{and} release of $\tau_{\text{ext}}$, driving the steady-state error to
zero; (iii)~keeps the robot--environment contact force within the ISO/TS 15066
safety limit \emph{at every instant}; and (iv)~respects the physical limits of
the transmission (actuator force, fluid pressure, jerk), at an update rate
realizable on embedded hardware. Fixed-gain impedance cannot meet (i)--(ii)
together---stiffness trades against disturbance rejection---and post-hoc
saturation cannot guarantee (iii)--(iv); this motivates the predictive,
offset-free formulation developed below.

\begin{table}[t]
\caption{Principal Symbols}
\label{tab:nomen}
\centering
\footnotesize
\setlength{\tabcolsep}{4pt}
\begin{tabular}{ll}
\toprule
Symbol & Meaning \\
\midrule
$\theta_f,\ \theta_d$ & finger joint angle; reference trajectory \\
$e=\theta_d-\theta_f$ & tracking error ($x_e=[e,\dot e]^\top$) \\
$\tau_{\text{ext}},\ F_{\text{ext}}$ & external contact torque / force at the fingertip \\
$F=\Fff+\Fmpc$ & actuator command (feedforward + MPC) \\
$A_1,A_2$ & master / slave piston areas \\
$P_1,P_2$ & master / slave fluid pressures \\
$x_1,x_2$ & master / slave piston displacements \\
$J_f(\theta_f)$ & finger-mechanism Jacobian ($\dot x_2=J_f\dot\theta_f$) \\
$\Ieff,\ \beff$ & effective joint inertia / damping \\
$\Gamma_e=1/\Ieff$ & effective compliance \\
$d,\ \dhat$ & lumped disturbance and its Kalman estimate \\
$N,\ \Delta t$ & MPC horizon length; sample period \\
\bottomrule
\end{tabular}
\end{table}

This paper takes as its base the constant-$A_d$ offset-free Impedance MPC framework established for redundant-manipulator 
pHRI \cite{cao2026phri}---a two-layer design (feedforward reduction to a configuration-independent double integrator, plus 
a receding-horizon QP with an encoder-only Kalman disturbance state) whose stability, recursive feasibility, and input-to-state 
stability (ISS) are analyzed at the \emph{architecture} level---and develops it for dexterous hands. The present manuscript 
is a simulation and modeling study rather than a hardware report. Operating on the residual error dynamics after feedforward 
reduction, the architecture is independent of the actuation technology; the contributions here are the parts \emph{not} contained in \cite{cao2026phri}:

\textbf{Platform reduction and substitution map} (Sections~\ref{sec:tendon}--\ref{sec:mpc}): an algebraic feedforward cancels 
the transmission's reflected inertia and damping---derived in detail for the hydraulic example, with the substitutions for the other tendon architectures in Table~\ref{tab:tendon}---reducing every platform to the constant-$A_d$
double integrator of \cite{cao2026phri} and enabling 500\,Hz operation with a precomputed QP cost inverse.

\textbf{Optional platform contact channel} (Section~\ref{sec:kalman}): the encoder-only offset-free Kalman estimator is 
inherited unchanged; where slave-piston pressure $P_2$ is instrumented, a \emph{sensorless} contact-torque estimate 
$\hat{\tau}_{\text{ext}} = A_2 P_2 J_f$ becomes an optional third measurement that shortens the contact-onset transient---not 
required for the offset-free guarantee and not used in the reported results.

\textbf{Platform constraint embedding with recursive feasibility} (Section~\ref{sec:hydraulic}): the actuation platform's 
physical limits enter the QP as constraints rather than post-hoc saturation---on the hydraulic example, cavitation, seal 
pressure, and the ISO/TS 15066 contact-force limit---with a soft-constraint relaxation that keeps the QP feasible under rigid impact.

\textbf{Benchmarks with independently verified realized impedance} (Sections~\ref{sec:sim1}--\ref{sec:sim2}): seven-controller 
studies on sinusoidal tracking and precision reach-and-hold identify the 500\,Hz update rate as the minimum requirement, for this 
plant and tuning, to suppress both contact-onset and release transients below a 15\,mrad criterion; the realized first-move stiffness 
driving the no-Kalman results is verified by three independent routes (Section~\ref{sec:stiffverify}).

\textbf{16-DOF simulation extension with a diagnosis and fix for the full-hand Kalman} (Section~\ref{sec:leap}): per-joint Impedance MPC 
scales to a LEAP Hand MuJoCo model in four grasping tasks; the three reasons the na\"ive per-joint Kalman fails on the coupled, 
contacting thumb---inertia mismatch, off-diagonal coupling, grasp self-contact---are isolated and removed in turn by gain scheduling, 
a coupled block estimator, and contact-aware conditional integration (thumb deviation $2329 \to 650 \to 236 \to 133$\,mrad, the 
last \emph{below} the 203\,mrad no-Kalman baseline).

\section{Related Work}
\label{sec:related}

\emph{Impedance MPC.} Cao \emph{et al.} \cite{cao2024passive} optimize impedance \emph{parameters} $\{M_d, D_d\}$ over a receding horizon 
with energy-tank passivity; because the parameters enter the prediction model nonlinearly, an iterative solver limits updates to 10--30\,Hz. 
Here the decision variables are corrective \emph{torques} and the constant $A_d$ permits offline $H^{-1}$ precomputation---a 15--50$\times$ 
update-rate advantage---while the energy-tank constraint of Section~\ref{sec:hydraulic} is directly inspired by \cite{cao2024passive}. 
Haninger \emph{et al.} \cite{haninger2023} jointly optimize force reference and impedance with Gaussian-process task models and probabilistic 
chance constraints; the present work instead uses a two-parameter analytic plant model, a hard deterministic ISO/TS 15066 bound, and an 
integrating disturbance state that achieves exactly zero steady-state error. Roveda \emph{et al.} \cite{roveda2020} adapt impedance parameters 
by model-based reinforcement learning, without per-step hard contact constraints or an offset-free estimator. Wang \emph{et al.} \cite{wu2025}
combine a linear MPIC with ADRC at 500\,Hz, but the ESO compensates the disturbance only at the current step; here the estimate is propagated 
through all $N$ prediction steps, which is why the controller pre-loads corrective torque and suppresses the contact-release transient to 
7.3\,mrad where reactive compensation retains 150\,mrad peaks.

\emph{Dexterous hands.} The LEAP Hand \cite{shaw2023leap} trades accuracy for safety with intentionally compliant joints 
($K_p = 3$\,Nm/rad)---precisely the trade-off Impedance MPC removes by enforcing contact force as a constraint independent of tracking 
stiffness; its MuJoCo model is the baseline throughout. RL-based in-hand manipulation \cite{andrychowicz2020} achieves impressive policies 
without formal guarantees on contact force or steady-state error, and safe contact reaction remains an open challenge \cite{deluca2006}. 
Hogan's formulation \cite{hogan1985}, the force-control survey \cite{villani2016}, and Merritt's hydraulic text \cite{merritt1967} underpin 
Section~\ref{sec:tendon}.

\section{Tendon Dynamics}
\label{sec:tendon}

The controller of Section~\ref{sec:mpc} requires from the transmission only two scalar parameters---the effective joint inertia $\Ieff$ and damping $\beff$---which exist for every tendon architecture. This section derives them for the hydraulic example and summarizes the substitutions (Table~\ref{tab:tendon}).

\subsection{Hydraulic Tendon (Example Platform)}

\subsubsection{Lumped Model and Joint-Space Reduction}

The motor--piston--fluid--finger transmission is modeled as a lumped second-order plant in the incompressible limit \cite{merritt1967}:
\begin{equation}
M_e\,\ddot{x}_2 + B_e\,\dot{x}_2 = \frac{A_2}{A_1}\,F - F_{\text{ext}}
\label{eq:lumped}
\end{equation}
where $x_2$ is the slave-piston displacement, $M_e = M_2 + M_1(A_2/A_1)^2$ and $B_e = b_2 + b_1(A_2/A_1)^2$ are the effective piston inertia and damping (master quantities reflected to the slave), $F$ is the motor drive force (control input), $F_{\text{ext}}$ the external contact force at the slave piston, and $A_1, A_2$ the piston areas. The finger linkage maps piston to joint motion through a scalar mechanism Jacobian $J_f(\theta_f)$ ($\dot{x}_2 = J_f\dot{\theta}_f$; $J_f = r_f$ constant for a linear tendon). Substituting and applying the virtual-work identity $\tau = J_f F$ yields the joint-space equation of motion
\begin{equation}
\Ieff\,\ddot{\theta}_f + \beff\,\dot{\theta}_f = \frac{A_2}{A_1}\,F\,J_f - \tau_{\text{ext}}
\label{eq:jointspace}
\end{equation}
with $\Ieff = M_e J_f^2$, $\beff = B_e J_f^2$, and $\tau_{\text{ext}} = F_{\text{ext}} J_f$; the centripetal-like term $M_e J_f\dot J_f\dot\theta_f^2$ vanishes for a linear tendon and is absorbed into the lumped disturbance otherwise. The effective compliance $\Gamma_e = 1/\Ieff$ plays the role of the operational-space inertia inverse $\Lambda^{-1}(q)$ in the Cartesian formulation of \cite{cao2026phri}.

\subsubsection{Sensorless Contact-Torque Estimation}
\label{sec:sensorless}

No dedicated force sensor is assumed. The slave pressure $P_2$ provides a contact-force estimate from Newton's law at the slave piston:
\begin{equation}
\hat{F}_{\text{ext}} = A_2\,P_2 - M_2\,\ddot{x}_2 - b_2\,\dot{x}_2 \approx A_2\,P_2 \quad\text{(quasi-static)}
\label{eq:sensorless}
\end{equation}
and $\hat{\tau}_{\text{ext}} = \hat{F}_{\text{ext}}\,J_f(\theta_f)$ in joint space. Pascal's law ($P_1 A_1 = P_2 A_2$ at quasi-static steady state) provides a redundancy check.

\subsubsection{Hydraulic Resonance and Bandwidth}
\label{sec:resonance}

With finite bulk modulus $\beta$, the fluid column resonates at
\begin{equation}
\omega_h = \sqrt{\frac{\beta}{V_e}\!\left(\frac{A_1^2}{M_1}+\frac{A_2^2}{M_2}\right)}
\label{eq:resonance}
\end{equation}
$\approx 3900$\,rad/s ($\approx 620$\,Hz) at typical values ($\beta = 1.5\times10^9$\,Pa, $A = 100$\,mm$^2$, $V_e = 10$\,mL, $M = 0.1$\,kg)---well above the closed-loop bandwidth realized at the 100--500\,Hz QP rates targeted here (dominant pole ${\approx}615$\,rad/s at 500\,Hz, Section~\ref{sec:stiffverify}).

\emph{Modeling limitations.} The lumped plant \eqref{eq:lumped} omits the servo-valve square-root flow nonlinearity, the temperature dependence of viscosity and bulk modulus, and finite-bandwidth compressibility. The first two are slowly varying and absorbed by the disturbance state $\dhat$; the third is benign while $\omega_h \gg$ bandwidth, but rigid impacts that excite the compressibility mode fall outside the quasi-static pressure map of Section~\ref{sec:pressure} and require an augmented pressure state, left to future work; the present results are the rigid-fluid, in-bandwidth limit.

\subsection{Other Tendon Architectures}

Only the feedforward inversion, the values of $(\Ieff, \beff)$, and the optional contact-measurement channel change across platforms; the error dynamics \eqref{eq:errdyn}--\eqref{eq:discrete}, estimator \eqref{eq:augmented}--\eqref{eq:measurement}, and QP are untouched. \emph{Cable tendons} \cite{jacobsen1986}---the most common architecture (Allegro, LEAP \cite{shaw2023leap}, Shadow)---give $\Ieff = I_m r_f^2/r_m^2$ from motor and capstan geometry, with motor current or cable tension as the contact channel. \emph{Pneumatic McKibben muscles} \cite{chou1996} require inverting the nonlinear force--pressure--length map in the feedforward and are bandwidth-limited to 5--10\,Hz by air compressibility. \emph{Twisted-string actuators} \cite{gaponov2014} require the nonlinear twist kinematics in the feedforward and suffer preload-dependent stiffness and wear. \emph{Series-elastic actuators} \cite{pratt1995} provide the cleanest contact channel---spring deflection measures transmitted torque directly---at the lowest bandwidth. Table~\ref{tab:tendon} summarizes.

\begin{table*}[!t]
\caption{Tendon Architecture Comparison}
\label{tab:tendon}
\centering
\footnotesize
\begin{tabular}{lccll}
\toprule
Architecture & Peak force density & Control bandwidth & Contact channel (replacing $P_2$) & Main limitation \\
\midrule
Hydraulic (example platform) & High & 50--500\,Hz & Slave pressure $P_2$ & Compressibility resonance; sealing \\
Cable tendon & Medium & 100--1000\,Hz & Motor current $i_m$ or cable tension & Back-drivability; cable stretch \\
Pneumatic PAM & Medium & 5--10\,Hz & PAM gauge pressure & Low bandwidth; nonlinear map \\
Twisted-string & High & 20--100\,Hz & Motor current or string tension & Nonlinear kinematics; wear \\
Series-elastic & Low--Medium & 10--30\,Hz & Spring deflection $\delta$ & Low bandwidth; added compliance \\
\bottomrule
\end{tabular}
\end{table*}

\section{Impedance MPC Design}
\label{sec:mpc}

\begin{figure}[t]
\centering
\resizebox{\columnwidth}{!}{%
\begin{tikzpicture}[every node/.style={font=\footnotesize}]
  \node (ref)        at (0,0)     {$\theta_d,\dot\theta_d,\ddot\theta_d$};
  \node[blk] (ff)    at (2.1,0)   {Feedforward\\ \eqref{eq:feedforward}};
  \node[sumc] (sum)  at (3.8,0)   {$+$};
  \node[blk] (plant) at (5.7,0)   {Finger\,+\\transmission};
  \node (out)        at (7.7,0)   {$\theta_f$};
  \node[circle,fill,inner sep=1pt] (fb) at (7.0,0) {};
  \node[sumc] (cmp)  at (1.0,-1.8) {};
  \node[est]  (est)  at (2.5,-1.8) {Kalman\\ estimator};
  \node[blk]  (qp)   at (5.6,-1.8) {Impedance MPC\\ QP \eqref{eq:condensed}};
  \node[draw,dashed,rounded corners=2pt,align=center,inner sep=2.5pt]
       (con)         at (5.6,-3.3) {constraints: ISO/TS\,15066,\\ pressure, motor, jerk};
  \draw[sig] (ref)--(ff);
  \draw[sig] (ff)-- node[above,font=\scriptsize]{$\Fff$} (sum);
  \draw[sig] (sum)-- node[above,font=\scriptsize]{$F$} (plant);
  \draw[sig] (plant)--(out);
  \draw[sig] (ref.south) |- (cmp.west);
  \node[font=\scriptsize] at (0.74,-1.6) {$+$};
  \draw[sig] (cmp)-- node[above,font=\scriptsize]{$e,\dot e$} (est);
  \draw[sig] (est)-- node[above,font=\scriptsize]{$\hat x_e,\dhat$} (qp);
  \draw[sig] (qp.north) -- ++(0,0.55) -|
        node[near end,left,font=\scriptsize]{$\Fmpc$} (sum.south);
  \draw[sig] (con)--(qp);
  \draw[sig] (fb) -- (7.0,-2.7) --
        node[below,pos=0.82,font=\scriptsize]{encoder $\theta_f$} (1.0,-2.7) -- (cmp.south);
  \node[font=\scriptsize] at (1.24,-2.02) {$-$};
\end{tikzpicture}}
\caption{Two-layer control architecture. Layer~1 (feedforward) cancels the
known transmission dynamics; Layer~2 (a 500\,Hz receding-horizon QP with an
encoder-only Kalman disturbance state $\dhat$) regulates the residual error
$e=\theta_d-\theta_f$ under hard contact-force, pressure, motor, and jerk
constraints.}
\label{fig:block}
\end{figure}
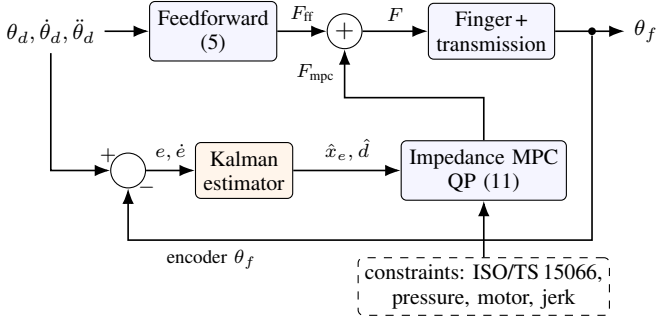

The two-layer architecture, force-form disturbance representation, offset-free Kalman estimator, and condensed-QP machinery are those of \cite{cao2026phri}, summarized here only to fix notation. New in this section are the hydraulic feedforward \eqref{eq:feedforward}, the optional pressure measurement channel, the first-move characterization of the realized impedance (Section~\ref{sec:equivalence}), and the explicit assumption-preservation argument under which the closed-loop guarantees of \cite{cao2026phri} transfer (Section~\ref{sec:inheritance}).

\emph{Command spaces.} Equation~\eqref{eq:jointspace} fixes a constant transmission gain $g_m \equiv (A_2/A_1)J_f$ relating motor force to joint torque, $\tau = g_m F$. The regulator is designed in \emph{joint-torque space}: the MPC decision variable is the joint torque $\tau_{\text{mpc}}$, the error dynamics \eqref{eq:errdyn} act on it through the joint compliance $\Gamma_e = 1/\Ieff$, and the realized first-move gain $k_e$ \eqref{eq:firstmove} is accordingly a joint stiffness (Nm/rad); this is the representation used by the released code. The actuator command is recovered by the inverse map $F = \tau/g_m$, and the platform limits of Section~\ref{sec:hydraulic} (motor force, piston pressure, ISO/TS~15066 contact force) are written in that force space. We keep the symbol $F_\bullet$ throughout: in the dynamics \eqref{eq:errdyn}, the gain \eqref{eq:firstmove}, and the estimator it denotes the joint torque $\tau_\bullet$, while in the feedforward \eqref{eq:feedforward} and the constraint rows \eqref{eq:constraints}--\eqref{eq:softqp} it denotes the motor-force image $\tau_\bullet/g_m$, the two related by the constant $g_m$ so the constant-$A_d$ structure is unaffected.

\subsection{Architecture and Error Dynamics}
\label{sec:errdynsec}

\textbf{Layer 1 --- Feedforward.} An algebraic term cancels the known transmission dynamics, leaving a scalar double integrator as the residual plant:
\begin{equation}
\Fff = \frac{A_1}{A_2\,J_f}\!\left[\Ieff\,\ddot{\theta}_d + \beff\,\dot{\theta}_f\right]
\label{eq:feedforward}
\end{equation}
The inertia term uses the reference acceleration $\ddot{\theta}_d$; the damping term uses the \emph{actual} velocity $\dot{\theta}_f$ so the physical damping cancels exactly. Only the two parameters $(\Ieff, \beff)$, identified from a step response, are required.

\textbf{Layer 2 --- Impedance MPC.} With $e = \theta_d - \theta_f$ and the joint-torque MPC correction $\Fmpc$ (\emph{Command spaces} above), the residual error dynamics are
\begin{equation}
\ddot{e} = -\Gamma_e(\theta_f)\,\Fmpc + d(t)
\label{eq:errdyn}
\end{equation}
where $d(t) = \tau_{\text{ext}}/\Ieff$ lumps contact torque and unmodeled terms. The state $x_e = [e,\dot e]^\top$ obeys $\dot x_e = A_c x_e + B_c \Fmpc + E_c d$ with $A_c$ the nilpotent double integrator, $B_c = [0;-\Gamma_e]$, $E_c = [0;1]$. As in \cite{cao2026phri}, the estimator and QP operate on the equivalent \emph{force-form} disturbance $d_F \equiv -\Ieff d$ ($= -\tau_{\text{ext}}$ for a pure contact torque), which enters through the same input matrix as the control and is the quantity denoted $\dhat$ below. Discretization is exact ($A_c^2 = 0$) and rate-independent in form:
\begin{equation}
A_d = \begin{bmatrix}1 & \Delta t \\ 0 & 1\end{bmatrix}, \qquad B_d = \begin{bmatrix}-\tfrac{\Delta t^2}{2}\Gamma_e \\ -\Delta t\,\Gamma_e\end{bmatrix}
\label{eq:discrete}
\end{equation}
The constancy of $A_d$ is the structural key inherited from \cite{cao2026phri}: the prediction matrices and the QP cost inverse $H^{-1}$ are computed once offline, reducing each MPC step to a matrix--vector multiply ($<0.1$\,ms on embedded hardware). For configuration-varying mechanisms (slider-crank linkages; the thumb axle of Section~\ref{sec:leap}, whose inertia varies 26$\times$), $A_d$ remains exactly constant and only $B_d(\theta_f)$---hence $\Gamma, H^{-1}$ and the Kalman input blocks---is rescheduled online, an $O(N^2)$ update per step.

\subsection{Offset-Free Disturbance Estimation}
\label{sec:kalman}

An integrating disturbance state is appended to the error state and estimated by a steady-state Kalman filter from \textbf{joint encoders alone} \cite{cao2026phri}, \cite{rawlings2017}, \cite{kalman1960}:
\begin{equation}
\begin{bmatrix} x_{e,k+1} \\ \hat{d}_{k+1} \end{bmatrix} = \begin{bmatrix} A_d & B_d \\ 0 & 1 \end{bmatrix} \begin{bmatrix} x_{e,k} \\ \hat{d}_k \end{bmatrix} + \begin{bmatrix} B_d \\ 0 \end{bmatrix} F_{mpc, k} + \begin{bmatrix} 0 \\ w_k \end{bmatrix}
\label{eq:augmented}
\end{equation}
\begin{equation}
y_k = \begin{bmatrix}1 & 0 & 0 \\ 0 & 1 & 0\end{bmatrix} [e,\;\dot{e},\;\dhat]^\top_k + v_k
\label{eq:measurement}
\end{equation}
where the zero-mean white process noise $w_k$, of variance $Q_{\text{proc},d}$, drives \emph{only} the disturbance state---the random-walk model $\dhat_{k+1} = \dhat_k + w_k$ (continuous time $\dot d = w$)---since the error states evolve deterministically once all unmodeled effects are lumped into $d$; $v_k$ is zero-mean white measurement noise with covariance $R_{\text{obs}}$. The random walk lets $\dhat$ track constant and slowly varying contact loads. Zero steady-state error under constant disturbances requires detectability of the augmented pair \cite{rawlings2017}, \cite{pannocchia2003}, which here reduces to the rank condition
\begin{equation}
\operatorname{rank}\begin{bmatrix} I - A_d & -B_d \\ C_{xe} & 0 \end{bmatrix} = n_x + n_d
\label{eq:rank}
\end{equation}
satisfied whenever $\Gamma_e > 0$. The estimate converges in 5--10 QP periods; the convergence rate---hence the contact-onset peak---is set by the noise ratio $Q_{\text{proc},d}/R_{\text{obs}}$ (Section~\ref{sec:sim1}). Unlike PI impedance, the structured observer gain prevents wind-up across sequential contact events \cite{cao2004antiwindup}, \cite{cao2002antiwindup}. A closed-loop analysis of the coupled observer--QP loop under \emph{active} input saturation remains open, as in \cite{cao2026phri}; offset-free behaviour is established in the nominal (constraint-inactive, detectable) limit and verified numerically.

\emph{Remark 1 (Estimator-agnostic offset-free property).} The offset-free property rests on the integrating internal model \eqref{eq:augmented}, not on the Kalman filter specifically: it holds for \emph{any asymptotically unbiased estimator of the augmented state satisfying the detectability condition \eqref{eq:rank}} \cite{rawlings2017}, \cite{pannocchia2003}---e.g.\ a momentum observer or an ESO in the ADRC sense \cite{wu2025}. The estimator choice sets only the transient convergence rate and noise response.

\emph{Optional platform contact channel.} Where slave pressure $P_2$ is instrumented (typically already, for safety interlocks), $y_3 = A_2 P_2 \approx -\dhat$ supplies a direct third measurement of the same force-form disturbance, shortening estimator lag from 5--10 QP periods to $\approx 1$. It is optional---the offset-free guarantee holds from encoders alone---and is \emph{not} exercised in the reported simulations; ``sensorless'' in Section~\ref{sec:sensorless} means no force/torque transducer. On other platforms the channel is replaced per Table~\ref{tab:tendon}.

\subsection{Receding-Horizon QP}
\label{sec:qp}

Stacking $N$ predicted states gives $Y = \Phi x_0 + \Gamma U + \bar D\dhat$, with $\Phi, \Gamma$ the standard prediction matrices built from \eqref{eq:discrete} and $\bar D_k = \sum_{l=0}^{k-1} A_d^{l} B_d$ the cumulative disturbance propagator of \cite{cao2026phri} (the force-form $\dhat$ enters through the same $B_d$ as the control).\footnote{The released implementation uses the equivalent acceleration-form disturbance $d = \tau_{\text{ext}}/\Ieff$ with input $G_d = [\Delta t^2/2,\ \Delta t]^\top = -\Ieff B_d$; since $G_d\,d = B_d\,d_F$, the predictor, the augmented model \eqref{eq:augmented}, and the offset-free property are identical.} Condensing the stage cost $\sum_k x_e^\top Q x_e + R_u \Fmpc^2$ (terminal weight $Q_f$) yields
\begin{equation}
H = \Gamma^\top\bar{Q}\Gamma + R_u I_N, \quad f = \Gamma^\top\bar{Q}\bigl(\Phi x_0 + \bar D\dhat\bigr)
\label{eq:condensed}
\end{equation}
and the QP $\min_U \tfrac12 U^\top H U + f^\top U$ subject to
\begin{equation}
\begin{aligned}
\Fffk{k} + \Fmpck{k} &\in [F_{\text{min}}, F_{\text{max}}] \;\;\text{(motor bounds)} \\
\tfrac{A_2}{A_1}\bigl(\Fffk{k} + \Fmpck{k}\bigr) &\leq 140\,\text{N} \;\;\text{(ISO/TS 15066)}\\
\bigl|\Delta\bigl(\Fffk{k} + \Fmpck{k}\bigr)\bigr| &\leq \Delta F_{\text{max}} \;\;\text{(jerk limit)}\\
P_{\text{vap}} \leq P_{2,k} &\leq P_{\text{max}} \;\;\text{(cavitation/seal)}
\end{aligned}
\label{eq:constraints}
\end{equation}
Only $\Fmpck{0}$ is applied. The optional energy-tank passivity constraint of Section~\ref{sec:hydraulic}-B, when enabled, adds one further row to \eqref{eq:constraints}. When no constraint is active the solve is the single multiply $U^* = -H^{-1}f$ with the offline $H^{-1}$, and otherwise OSQP \cite{osqp} or qpOASES \cite{qpoases} solves the small ($N{=}10$) QP warm-started in $<0.05$\,ms on a Raspberry Pi~4. The constant $A_d$ is also what collapses the min--max LPV saturation formulation of \cite{cao2005minmax} to a single nominal QP.

\subsection{Equivalence to Classical Impedance and Realized Stiffness}
\label{sec:equivalence}

In the unconstrained, disturbance-free, infinite-horizon limit the Impedance MPC is a static LQR feedback rendering the classical impedance law $\tau^{\text{cmd}} = \Ieff\ddot\theta_d + \beff\dot\theta_d + K_{\text{eff}} e + D_{\text{eff}} \dot e$, where the realized gains $(K_{\text{eff}},D_{\text{eff}})$ are the LQR (Riccati) image of the weights $(Q,R)$---in general not the weights themselves, and not the cheap-control limit $R_u\to0$. The weights are the design handle ($Q_{\text{pos}} = K_d$, $Q_{\text{vel}} = D_d$); at finite horizon and sampling rate the realized impedance differs further from this nominal target, which we characterize by the \emph{first-move gain}
\begin{equation}
[k_e \;\; d_e] = -\bigl(H^{-1}\Gamma^\top\bar{Q}\,\Phi\bigr)_{1,:}
\label{eq:firstmove}
\end{equation}
so that the applied input is $\Fmpck{0} = k_e e + d_e \dot e$ when unconstrained. For the horizon, weights, and plant used in this paper, $k_e$ increases with the QP update rate and saturates in $Q_{\text{pos}}$; this monotonic rate dependence is a property of the specific tuning, established by direct computation and pole analysis in Section~\ref{sec:stiffverify}---not claimed as universal.

\subsection{Stability, Recursive Feasibility, and ISS: Inheritance}
\label{sec:inheritance}

No closed-loop guarantees are re-derived in this paper. The base framework \cite{cao2026phri} establishes, on the reduced system \eqref{eq:errdyn}--\eqref{eq:augmented}, (i) nominal asymptotic stability under the terminal Riccati cost with sufficient horizon, (ii) recursive feasibility via soft-constraint relaxation, and (iii) an ISS bound with respect to the estimation error $d - \dhat$. That analysis is embodiment-independent provided the following are preserved:
\begin{itemize}
\item[\textbf{A1}] \emph{Bounded-error cancellation:} the feedforward \eqref{eq:feedforward} reduces the plant to \eqref{eq:errdyn} with residual mismatch absorbed into a bounded $d(t)$.
\item[\textbf{A2}] \emph{Stabilizable residual dynamics:} the double integrator \eqref{eq:errdyn} with $\Gamma_e > 0$, or its block-diagonal multi-finger extension.
\item[\textbf{A3}] \emph{Positive-definite cost:} $Q \succ 0$, $R_u > 0$, $Q_f \succeq Q$.
\item[\textbf{A4}] \emph{Convex constraints:} all rows of \eqref{eq:constraints} linear in $U$, with the physical limits slack-relaxed (Section~\ref{sec:pressure}) so the feasible set is never empty.
\end{itemize}
The hand embodiments preserve A1--A4 in the reduced residual model. In particular, the hydraulic pressure limits enter as \emph{quasi-static output constraints} on the same residual dynamics---pressure is \emph{not} a prediction-model state---so the nominal closed-loop structure matches the one analyzed in \cite{cao2026phri}. We therefore use those results as an architectural guarantee under the stated assumptions rather than as a new proof for every contact-rich hand configuration. By Remark~1 the offset-free property likewise applies under any detectable, asymptotically unbiased estimator in the nominal constraint-inactive limit. Two boundaries are explicit: active saturation/contact switching is validated numerically rather than re-proved here, and in the compressibility regime of Section~\ref{sec:resonance}, $P_2$ must become a prediction state, the augmented model differs from the one analyzed in \cite{cao2026phri}, and the inherited guarantees would require re-verification; that regime is excluded here.

\section{Platform Constraints: The Hydraulic Example}
\label{sec:hydraulic}

The constraint-embedding pattern is platform-independent---any actuation technology contributes its physical limits as linear (or linearized quasi-static) inequalities on $U$, e.g.\ cable-tension positivity, PAM pressure bounds, SEA deflection limits. This section instantiates it on the hydraulic example.

\subsection{Pressure Constraint and Recursive Feasibility}
\label{sec:pressure}

The seal limit $P_{2,k} \leq P_{\text{max}}$ maps to a linear inequality on the command:
\begin{equation}
\Fffk{k} + \Fmpck{k} \leq \frac{A_1}{A_2}\!\left[A_2\,P_{\text{max}} - \frac{\Ieff}{J_f}\dhat_k - B_e\,\dot{x}_{2,k}\right]
\label{eq:pressureconstraint}
\end{equation}
(the factor $\Ieff/J_f$ converts the acceleration-form estimate $\dhat_k = \hat\tau_{\text{ext},k}/\Ieff$ of Section~\ref{sec:qp} to the piston-space contact force $\hat F_{\text{ext},k} = \hat\tau_{\text{ext},k}/J_f$), and likewise for cavitation $P_{2,k} \geq P_{\text{vap}}$. Enforcing these inside the QP guarantees satisfaction at every predicted step---unlike post-hoc saturation, which alters the applied torque after the solve and violates passivity \cite{focchi2012}. Two caveats: (i) the map is \emph{quasi-static}---$P_2$ is an algebraic output of the force balance, not a state---justified by the timescale separation of Section~\ref{sec:resonance} (the fluid mode at $\omega_h \approx 3900$\,rad/s lies a factor ${\approx}6$ above the realized closed-loop bandwidth, so the pressure transient decays within each control interval). When that separation fails, $P_2$ must be predicted as a state, with the consequence noted in Section~\ref{sec:inheritance}. (ii) Hard physical limits can render the QP infeasible under rigid impact---if $\Fff$ alone exceeds the pressure budget, no admissible $\Fmpc$ exists. The physical limits are therefore relaxed to penalized soft constraints,
\begin{align}
\min_{U,s}\ &\tfrac12 U^\top H U + f^\top U + \rho\,\textstyle\sum_k s_k^2, \nonumber\\
&\tfrac{A_2}{A_1}\big(\Fffk{k}+\Fmpck{k}\big) \le F_{\text{contact,max}} + s_k,\quad s_k \ge 0
\label{eq:softqp}
\end{align}
(and likewise for the pressure rows), keeping the torque/jerk box hard. This preserves the constant-$A_d$ structure and guarantees recursive feasibility: the QP returns the \emph{minimum-violation} command rather than faulting. Under an impact feedforward that empties the hard set, the hard QP reports primal infeasible while the slacked QP returns the motor-saturated command with $s$ equal to the unavoidable over-pressure---the same device used for the torque limit in \cite{cao2026phri}, maintaining assumption A4.

\subsection{Further Extensions}

\emph{Passivity.} An energy-tank constraint $T_{k+1} = T_k + \tau_{\text{ext},k}\dot\theta_{f,k}\Delta t - F_{\text{mpc},k}J_f\dot\theta_{f,k}\Delta t \geq 0$, linear in $\Fmpc$, bounds the energy injected into the environment, extending \cite{ott2008}, \cite{cao2024passive} to the receding-horizon setting. \emph{Variable impedance.} Task-dependent compliance is obtained by scheduling $Q(t) = \text{diag}(K_d(t), D_d(t))$, e.g.\ lowering $K_d$ when $\|\hat\tau_{\text{ext}}\|$ exceeds a contact threshold; the QP re-solves from the current state with no discontinuity. \emph{Multi-finger.} Each finger runs an independent scalar QP; object-level grasp coordination adds a wrench condition coupling them into an $nN$-variable QP, solvable in $<1$\,ms for $n=5$, $N=10$.

\section{Simulation Study: Disturbance Rejection Under Sinusoidal Trajectory}
\label{sec:sim1}

\subsection{Scenario and Controllers}
\label{sec:controllers}

Seven controllers (Table~\ref{tab:controllers}) track $\theta_d(t) = 0.8 + r(t)\,0.4\sin(\tfrac{\pi}{2} t)$\,rad ($r(t) = \min(1, t)$, smooth start) while a step contact torque $\tau_{\text{ext}} = 1.5$\,Nm is applied from $t = 1.5$ to $3.0$\,s of each 4\,s cycle and released; four cycles are simulated. Plant: $I_{\text{eff}} = 10^{-3}$\,kg$\cdot$m$^2$, $b_{\text{eff}} = 2\times10^{-3}$\,Nm$\cdot$s/rad, $\Gamma_e = 1/\Ieff = 10^3\ (\text{kg}\cdot\text{m}^2)^{-1}$; the normalized disturbance $d = 1500$\,rad/s$^2$ is large relative to the torque limit. All MPC variants use $N = 10$, $Q_{\text{pos}} = 10^8$, $Q_{\text{vel}} = 30$, $Q_{f,\text{scale}} = 5$, $R_u = 10^{-6}$, $\tau_{\text{max}} = 3$\,Nm; D4/D5 run at 100\,Hz, D6/D7 at 500\,Hz, D5/D7 with the Kalman estimator.

The weight $Q_{\text{pos}}$ is set near the sampling-rate stiffness ceiling: for this tuning the realized first-move stiffness \eqref{eq:firstmove} saturates at $k_e\approx 18$\,Nm/rad at 100\,Hz and reaches $k_e\approx 323$\,Nm/rad at 500\,Hz (verified in Section~\ref{sec:stiffverify}), both above the classical $K_d = 10$. Pushing $Q_{\text{pos}}$ higher only saturates $k_e$ (the 100\,Hz ceiling is $\approx 20$, beyond which the discrete loop overshoots) while growing the Hessian conditioning (measured $\operatorname{cond}(H) \approx 1.3\times10^{6}$ at 100\,Hz, $1.1\times10^{5}$ at 500\,Hz). The torque box is inactive in these tracking tasks, so the analytic $H^{-1}$ solve is exact.

\begin{table}[!t]
\caption{Controllers Evaluated in Simulation}
\label{tab:controllers}
\centering
\footnotesize
\setlength{\tabcolsep}{2.5pt}
\begin{tabular}{clccl}
\toprule
ID & Controller & Rate & Dist.\ est. & Paradigm \\
\midrule
D1 & Classical impedance ($K_d{=}10$) & 1\,kHz & None & Reactive \\
D2 & Admittance ($K_a{=}3$) & 1\,kHz & None & Yield to contact \\
D3 & PI impedance ($K_{\text{int}}{=}4$) & 1\,kHz & Integral & React.\ + integral \\
D4 & Impedance MPC & 100\,Hz & None & Predictive \\
D5 & Impedance MPC + Kalman & 100\,Hz & Kalman $\dhat$ & Pred.\ + offset-free \\
D6 & Impedance MPC & 500\,Hz & None & Predictive, fast \\
D7 & Impedance MPC + Kalman & 500\,Hz & Kalman $\dhat$ & Pred.\ + offset-free \\
\bottomrule
\end{tabular}
\end{table}

\subsection{Results and Analysis}

\begin{figure}[!t]
\centering
\includegraphics[width=0.78\columnwidth]{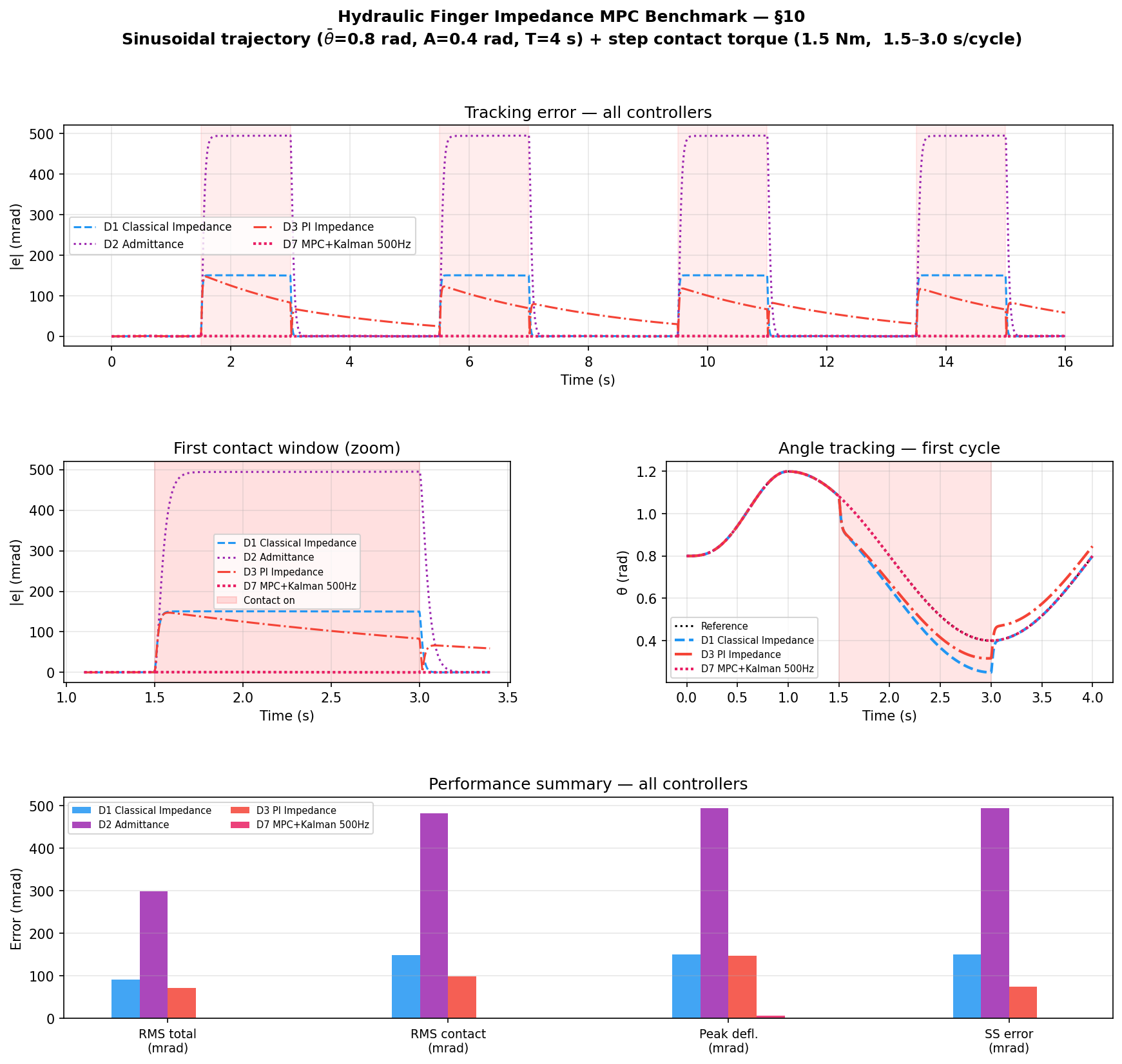}
\caption{Sinusoidal tracking under cyclic 1.5\,Nm step contact torque: the classical baselines (D1--D3) versus the proposed 500\,Hz Kalman MPC (D7). D7's error (7.3\,mrad peak, 0.1\,mrad steady-state) is barely distinguishable from zero at this scale; the remaining MPC variants are omitted for readability and reported in Table~\ref{tab:sinresults}.}
\label{fig:sinusoidal}
\end{figure}

Table~\ref{tab:sinresults} reports RMS error over the full run, RMS and peak deflection within the force-on windows, and steady-state (SS) error at the end of each contact window; Fig.~\ref{fig:sinusoidal} shows the traces.

\begin{table}[!t]
\caption{Sinusoidal Tracking Results ($\tau_{\text{ext}} = 1.5$\,Nm, 16\,s / 4 Cycles)}
\label{tab:sinresults}
\centering
\footnotesize
\setlength{\tabcolsep}{4pt}
\begin{tabular}{lcccc}
\toprule
Controller & \shortstack{RMS tot.\\(mrad)} & \shortstack{RMS cont.\\(mrad)} & \shortstack{Peak\\(mrad)} & \shortstack{SS\\(mrad)} \\
\midrule
D1 --- Impedance & 91.7 & 149.1 & 150.6 & 150.1 \\
D2 --- Admittance & 299.0 & 482.9 & 495.0 & 494.9 \\
D3 --- PI Impedance & 72.3 & 98.8 & 147.3 & 74.4 \\
D4 --- MPC (no Kalman) & 52.1 & 84.5 & 150.5 & 83.3 \\
D5 --- MPC + Kalman 100\,Hz & 10.3 & 15.3 & 150.6 & \textbf{0.1} \\
D6 --- MPC 500\,Hz & 3.1 & 5.1 & 6.4 & 4.7 \\
D7 --- MPC + Kalman 500\,Hz & \textbf{0.6} & \textbf{0.7} & \textbf{7.3} & \textbf{0.1} \\
\bottomrule
\end{tabular}
\end{table}

\emph{Baselines match theory.} D1 reaches 150.1\,mrad SS, matching $\tau_{\text{ext}}/K_d = 150$\,mrad and validating the plant; D2 yields $\approx \tau_{\text{ext}}/K_a = 500$\,mrad by design; D3's integral action converges too slowly within the 1.5\,s contact window under its anti-windup limit (74.4\,mrad).

\emph{No-Kalman MPC is stiffness-limited but already better than classical.} With no disturbance state the QP is a predictive realization of an impedance with $e_\infty = \tau_{\text{ext}}/k_e$. At 100\,Hz, $k_e \approx 18$ halves D1's error (83.3\,mrad SS); at 500\,Hz, $k_e \approx 323$ reaches 4.7\,mrad SS---a 32$\times$ improvement with no estimator. Beyond the weight ceiling the \emph{rate}, not the weight, is the lever. Two limits remain: a residual offset, and---at 100\,Hz---a contact-onset peak (150.5\,mrad) set by the 10\,ms ZOH window that no weight choice removes (D6 peak: 6.4\,mrad).

\emph{The Kalman estimator removes the offset; the rate governs the transient.} Adding the estimator drives SS error to 0.1\,mrad at both rates (833$\times$ at 100\,Hz, 47$\times$ at 500\,Hz): once $\dhat$ converges, the disturbance is pre-canceled through all $N$ prediction steps. Convergence takes 5--10 QP periods, so at 100\,Hz the onset peak still builds to D1 levels (D5: 150.6\,mrad) while at 500\,Hz it is contained (D7: 7.3\,mrad, 21$\times$ better than D1). The noise ratio $Q_{\text{proc},d}/R_{\text{obs}}$ sets the convergence speed. D7 is best on all four metrics; the rate and estimator contributions are orthogonal (transient vs.\ steady state) and additive. For reference, the LEAP Hand's compliant design ($K_p = 3$\,Nm/rad) would deflect 500\,mrad under this load; D7 holds 0.1\,mrad while enforcing the ISO/TS 15066 limit as a hard constraint.

\subsection{Independent Verification of the Realized Stiffness}
\label{sec:stiffverify}

The 18$\to$323\,Nm/rad stiffness increase from a 5$\times$ rate change is the largest quantitative claim of the no-Kalman results and is verified by three routes independent of the benchmark code path (script in the supplementary material). \emph{(1) Gain extraction:} building $H, \Gamma, \bar Q, \Phi$ directly from \eqref{eq:discrete}, \eqref{eq:condensed} and evaluating \eqref{eq:firstmove} gives $k_e = 18.02$, $d_e = 0.190$ at 100\,Hz and $k_e = 323.1$, $d_e = 0.823$ at 500\,Hz. \emph{(2) Closed-loop poles:} the eigenvalues of $A_d + B_d[k_e\;d_e]$ are $\{0, -0.802\}$ at 100\,Hz and $\{0, -0.292\}$ at 500\,Hz---Schur-stable, real spectra, the shrinking dominant pole confirming the faster loop tolerates the larger gain. \emph{(3) Steady-state prediction:} $e_\infty = \tau_{\text{ext}}/k_e$ predicts 83.2 and 4.6\,mrad; the closed-loop iteration reproduces these to three digits, and the full benchmark measures 83.3 (D4) and 4.7\,mrad (D6)---agreement within 0.1\,mrad. The mechanism is the sampled-data realization: with $Q_{\text{pos}}$ saturated, the admissible first-move gain is bounded by the ZOH interval on this low-inertia plant, and shrinking $\Delta t_{\text{MPC}}$ relaxes the bound; the claim holds \emph{for the chosen horizon, weights, and plant}, with no universal monotonicity across tunings asserted.

\section{Simulation Study: Precision Reach-and-Hold Under Object Contact}
\label{sec:sim2}

\subsection{Scenario}

Three sequential waypoints are commanded (A: 0.5\,rad / 1.5\,Nm; B: 1.0\,rad / 2.0\,Nm, both opposing flexion; C: 1.4\,rad / 1.0\,Nm into the joint stop), the contact firing 0.5\,s after waypoint arrival. The \emph{advance condition}---the precision requirement for dexterous grasping---is that the finger remain within 15\,mrad of the waypoint for 0.5\,s after the contact window closes.

\subsection{Results and Analysis}

\begin{table*}[!t]
\caption{Precision Reach-and-Hold Results (Advance Criterion: $|e| \leq 15$\,mrad)}
\label{tab:graspresults}
\centering
\footnotesize
\begin{tabular}{lccccc}
\toprule
Controller & WP passed & RMS approach (mrad) & RMS contact (mrad) & Peak defl.\ (mrad) & SS @A/@B/@C (mrad) \\
\midrule
G1 --- Stiff Impedance & 0/3 & 0.3 & 154 & 200 & 150 / 200 / 100 \\
G2 --- Admittance & 0/3 & 0.3 & 501 & 660 & 495 / 660 / 330 \\
G3 --- PI Impedance & 0/3 & 45.7 & 118 & 171 & 87 / 101 / 80 \\
G4 --- MPC 100\,Hz & 0/3 & 3.2 & 89 & 295 & 83 / 111 / 56 \\
G5 --- MPC + Kalman 100\,Hz & 0/3 & 0.3 & 25 & 300 & 0 / 0 / 0 \\
G6 --- MPC 500\,Hz & \textbf{3/3} & 0.3 & 5 & 12 & 5 / 6 / 3 \\
\textbf{G7 --- MPC + Kalman 500\,Hz} & \textbf{3/3} & \textbf{0.3} & \textbf{0.5} & \textbf{12} & \textbf{0 / 0 / 0} \\
\bottomrule
\end{tabular}
\end{table*}

\begin{figure}[!t]
\centering
\includegraphics[width=0.78\columnwidth]{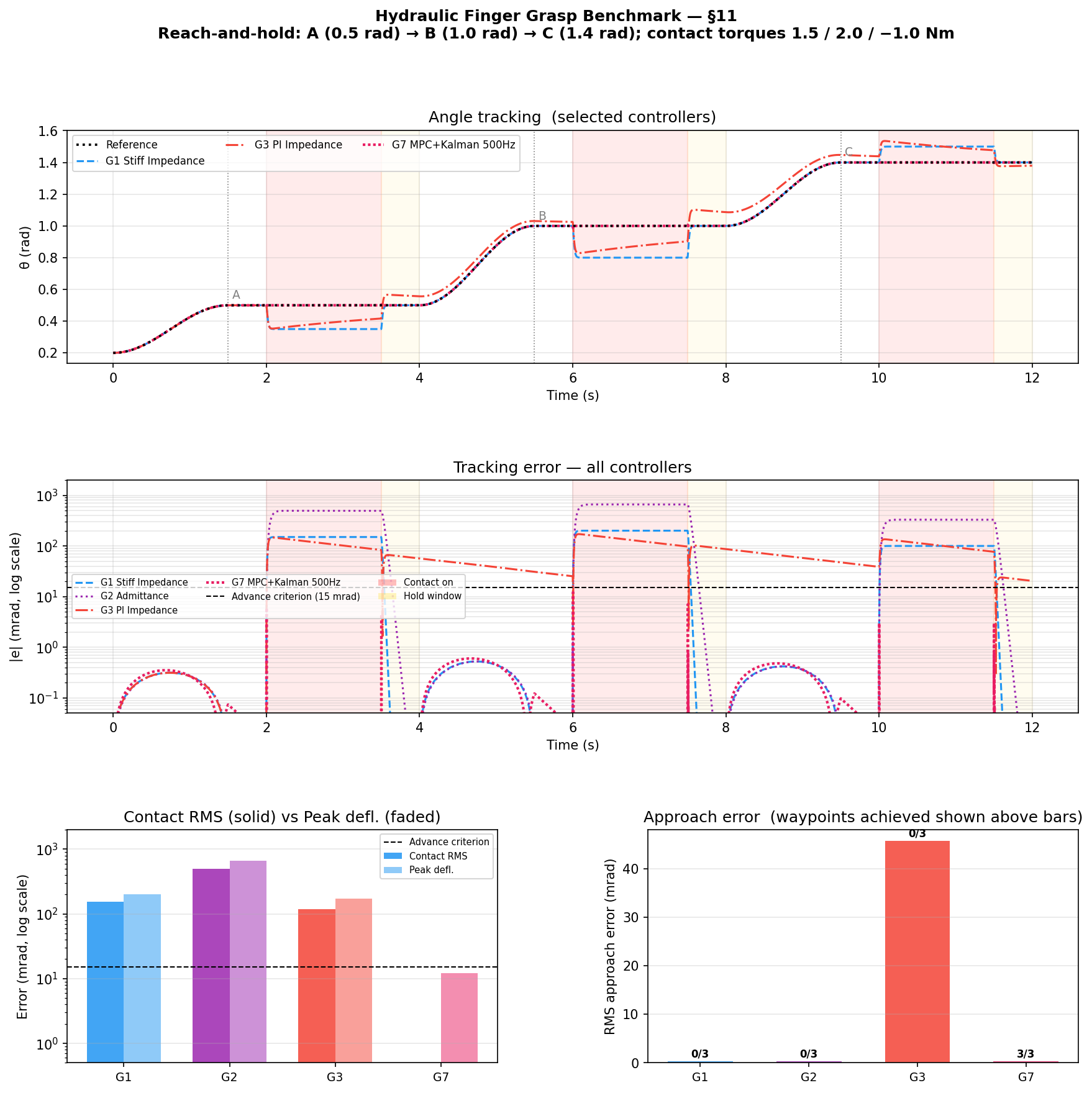}
\caption{Precision reach-and-hold benchmark: the classical baselines (G1--G3) violate the 15\,mrad advance criterion at every waypoint, while the proposed G7 passes all three with zero steady-state error. The remaining MPC variants are omitted for readability and reported in Table~\ref{tab:graspresults}.}
\label{fig:grasp}
\end{figure}

Table~\ref{tab:graspresults} and Fig.~\ref{fig:grasp} report the results.

\emph{100\,Hz fails on transients, not steady state.} G1's SS errors ($\tau_{\text{ext}}/K_d$: 150/200/100\,mrad) violate the criterion outright. G4 halves them ($k_e \approx 18$) yet is disqualified by a \emph{larger} onset peak (295 vs.\ 200\,mrad): over the 10\,ms ZOH window the 2\,Nm contact ($2000$\,rad/s$^2$) builds $\approx d\Delta t^2/2 = 100$\,mrad of error before the QP responds, and the ZOH-held damping cannot arrest the momentum without overshoot---no $(Q,R)$ choice removes this; only the rate does. G5 reaches 0\,mrad SS at every waypoint but fails on the \emph{release transient}: at contact release $\dhat$ still carries the previous load for 5--10 QP periods (50--100\,ms at 100\,Hz), overshooting to 300\,mrad---a failure mode invisible in the cyclic averages of Section~\ref{sec:sim1}. G3 additionally shows 45.7\,mrad approach error on the A$\to$B ramp from integral wind-up accumulated at A---cross-contamination between contact events that the structured observer avoids.

\emph{500\,Hz passes; the Kalman adds exactness.} G6 keeps every deflection within the criterion (peak 12\,mrad, SS 5/6/3\,mrad) and passes 3/3 with no estimator; G7 converges within 2--4 updates at onset \emph{and} release, holding the same 12\,mrad peak with exactly zero SS error. For this plant and tuning, the 500\,Hz rate is therefore necessary and sufficient to bring every contact transient within the precision window---at 100\,Hz both MPC variants fail, each on a different transient.

\section{16-DOF Extension: LEAP Hand MuJoCo Simulation}
\label{sec:leap}

\subsection{Setup and the Full-Hand Kalman}
\label{sec:leapkalman}

The single-finger controller extends to the 16-DOF LEAP Hand in MuJoCo \cite{todorov2012} by running an independent per-joint scalar QP at 500\,Hz, with per-joint $I_{\text{eff}}$ and $\tau_{\text{max}}$ from mass-matrix measurements (Table~\ref{tab:perjoint}), feedforward damping cancellation per \eqref{eq:feedforward}, and gravity compensation from the generalized bias force.

On the full hand the na\"ive per-joint Kalman fails on the thumb, for three separable reasons, each isolated and removed in turn:

\emph{(i) Inertia mismatch $\to$ gain scheduling.} The thumb-axle inertia varies 26$\times$ across the workspace; a fixed-$I_{\text{eff}}$ Kalman reads the mismatch as a persistent disturbance and chases it. Because $A_d$ never depends on $I_{\text{eff}}$, the inertia is gain-scheduled online from the mass-matrix diagonal, $I_{\text{eff}}(\theta) = M_{jj}(q)$, rebuilding only $B_d, \Gamma, H^{-1}$ and the Kalman input blocks (Section~\ref{sec:errdynsec}). A scalar study sweeping the full 26$\times$ range confirms: fixed-inertia RMS 1.2\,mrad with 57\,mrad/s jitter vs.\ gain-scheduled 0.4\,mrad and 3\,mrad/s---an ${\sim}18\times$ reduction.

\emph{(ii) Off-diagonal coupling $\to$ block estimator.} A per-joint (diagonal) estimator structurally cannot model the thumb's off-diagonal mass coupling and chases the coupling torques: under a 2.5\,N fingertip push at the power grasp, the per-joint Kalman reaches 2329\,mrad thumb deviation (sometimes divergent) vs.\ 203\,mrad with no Kalman; diagonal gain scheduling only partly helps (650\,mrad). The fix is a \emph{coupled block estimator} on the $4\times4$ thumb mass block $M_{\text{th}}(q)$: the residual is the block double integrator $\ddot e = -M_{\text{th}}^{-1}(q)\tau_{\text{mpc}} + d$ with a coupled $\dhat \in \mathbb{R}^4$. A 2-DOF study with strong coupling ($c/\sqrt{m_1 m_2} = 0.61$) isolates the mechanism: per-joint Kalman \emph{worse} than none (738 vs.\ 127\,mrad RMS), block Kalman best (38\,mrad). On the full hand the block estimator removes the instability: 236\,mrad deviation, no divergence.

\emph{(iii) Self-contact $\to$ contact-aware integration.} At curled grasp poses the thumb's self-contact forces are not in $M_{\text{th}}(q)$, so the integrating $\dhat$ winds up against the unrejectable contact reaction. \emph{Contact-aware conditional integration} suspends (leaks) the disturbance integration on joints registering a contact reaction and resumes the instant contact clears---the multi-DOF, contact-gated analogue of saturation anti-windup \cite{cao2004antiwindup}, \cite{cao2002antiwindup}. With the gate, the thumb deviation falls to 133\,mrad---\emph{below} the 203\,mrad no-Kalman baseline---and the gate is inert when the thumb is contact-free.

The progression $2329 \to 650 \to 236 \to 133$\,mrad attributes the failure to inertia mismatch, coupling, and contact in turn, and reconciles the full hand with the unambiguously beneficial single-finger Kalman: a single finger has neither coupling nor self-contact. The full-hand results below use the conservative no-Kalman default; the block + contact-aware estimator is the validated upgrade.

\begin{table}[!t]
\caption{Per-Joint MPC Parameters (LEAP Hand)}
\label{tab:perjoint}
\centering
\footnotesize
\setlength{\tabcolsep}{3pt}
\begin{tabular}{lccl}
\toprule
Joint type & $I_{\text{eff}}$ (kg$\cdot$m$^2$) & $\tau_{\text{max}}$ (Nm) & Rule \\
\midrule
Finger MCP & $4.74\times10^{-4}$ & 0.50 & $I_{\text{min}} \times 3000$\,rad/s$^2$, capped \\
Finger ROT & $5.46\times10^{-5}$ & 0.16 & $I_{\text{min}} \times 3000$\,rad/s$^2$ \\
Finger PIP & $9.73\times10^{-5}$ & 0.29 & $I_{\text{min}} \times 3000$\,rad/s$^2$ \\
Finger DIP & $1.47\times10^{-5}$ & 0.044 & $I_{\text{const}} \times 3000$\,rad/s$^2$ \\
Thumb CMC & $4.80\times10^{-4}$ & 0.048 & $I_{\text{avg}} \times 100$\,rad/s (coupling) \\
Thumb AXL & $9.90\times10^{-6}$ & 0.030 & $I_{\text{min}} \times 3000$\,rad/s$^2$; hard cap \\
Thumb MCP & $2.83\times10^{-4}$ & 0.028 & $I_{\text{avg}} \times 100$\,rad/s (coupling) \\
Thumb IPL & $4.95\times10^{-5}$ & 0.005 & $I_{\text{const}} \times 100$\,rad/s \\
\bottomrule
\end{tabular}
\end{table}

The thumb $\tau_{\text{max}}$ rules are deliberately tighter ($\times 100$\,rad/s vs.\ $\times 3000$\,rad/s$^2$): aggressive CMC correction couples ${\sim}3$\,Nm onto the thumb MCP through the off-diagonal mass matrix, and the tighter bound ($\Delta v/\text{step} \leq 0.2$\,rad/s) prevents the resulting limit cycle. The thumb-axle cap $\tau_{\text{max}} = I_{\text{min}}\times3000 = 0.030$\,Nm retains a 26$\times$ safety margin across its inertia variation pending the gain-scheduled estimate of Section~\ref{sec:leapkalman}.

\subsection{Results}

\begin{figure*}[!t]
\centering
\begin{tabular}{ccc}
\includegraphics[width=0.31\textwidth]{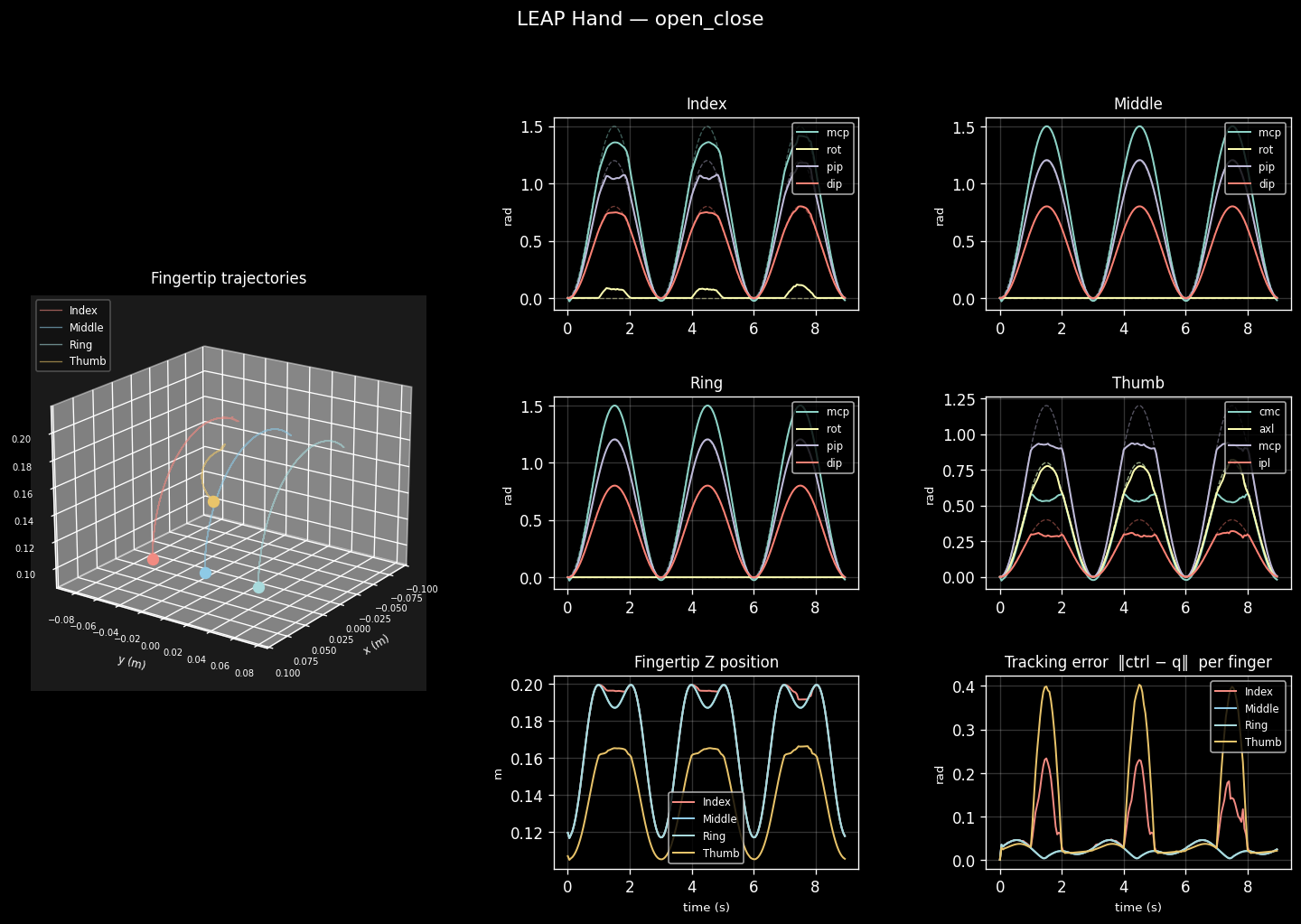} &
\includegraphics[width=0.31\textwidth]{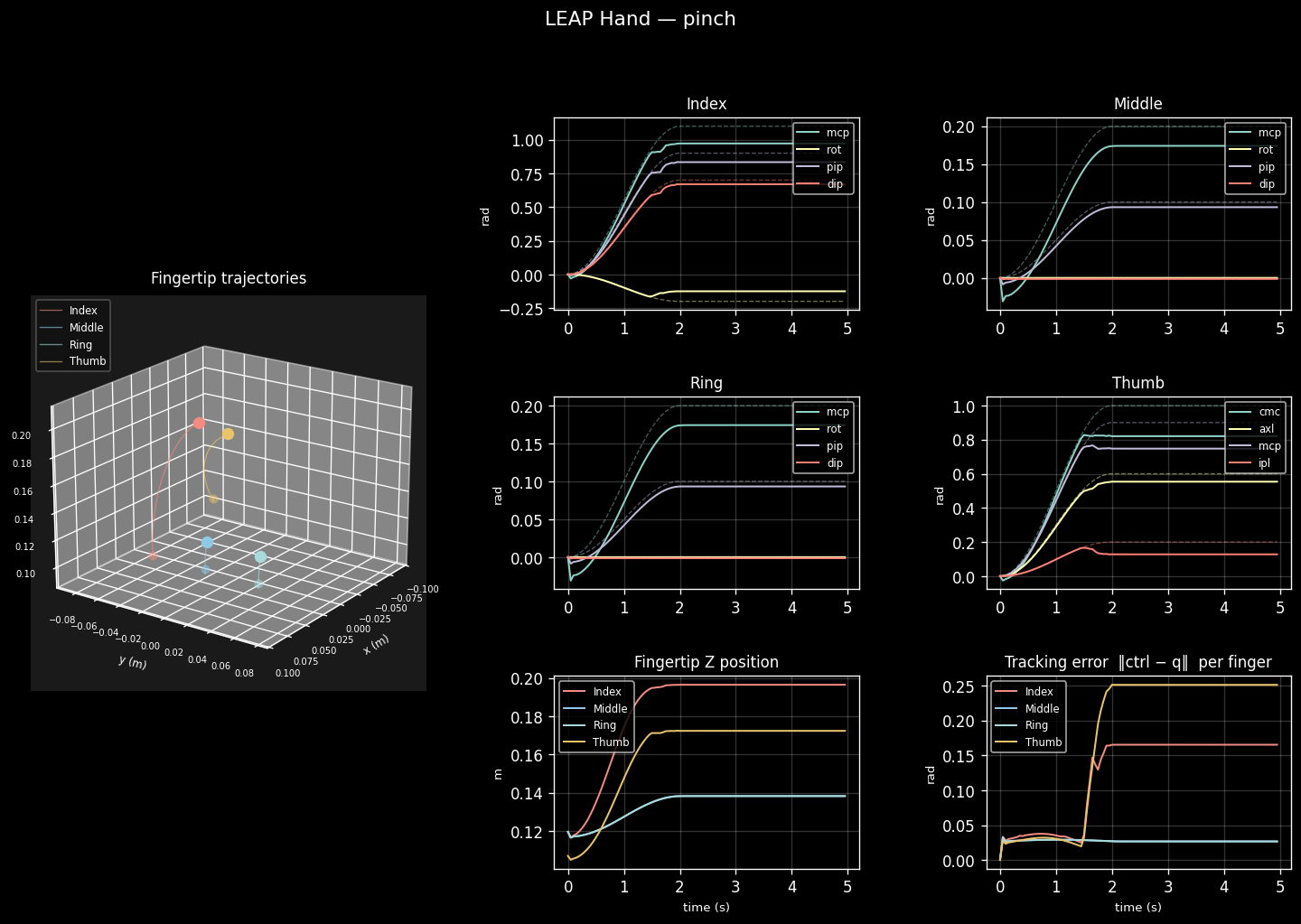} &
\includegraphics[width=0.31\textwidth]{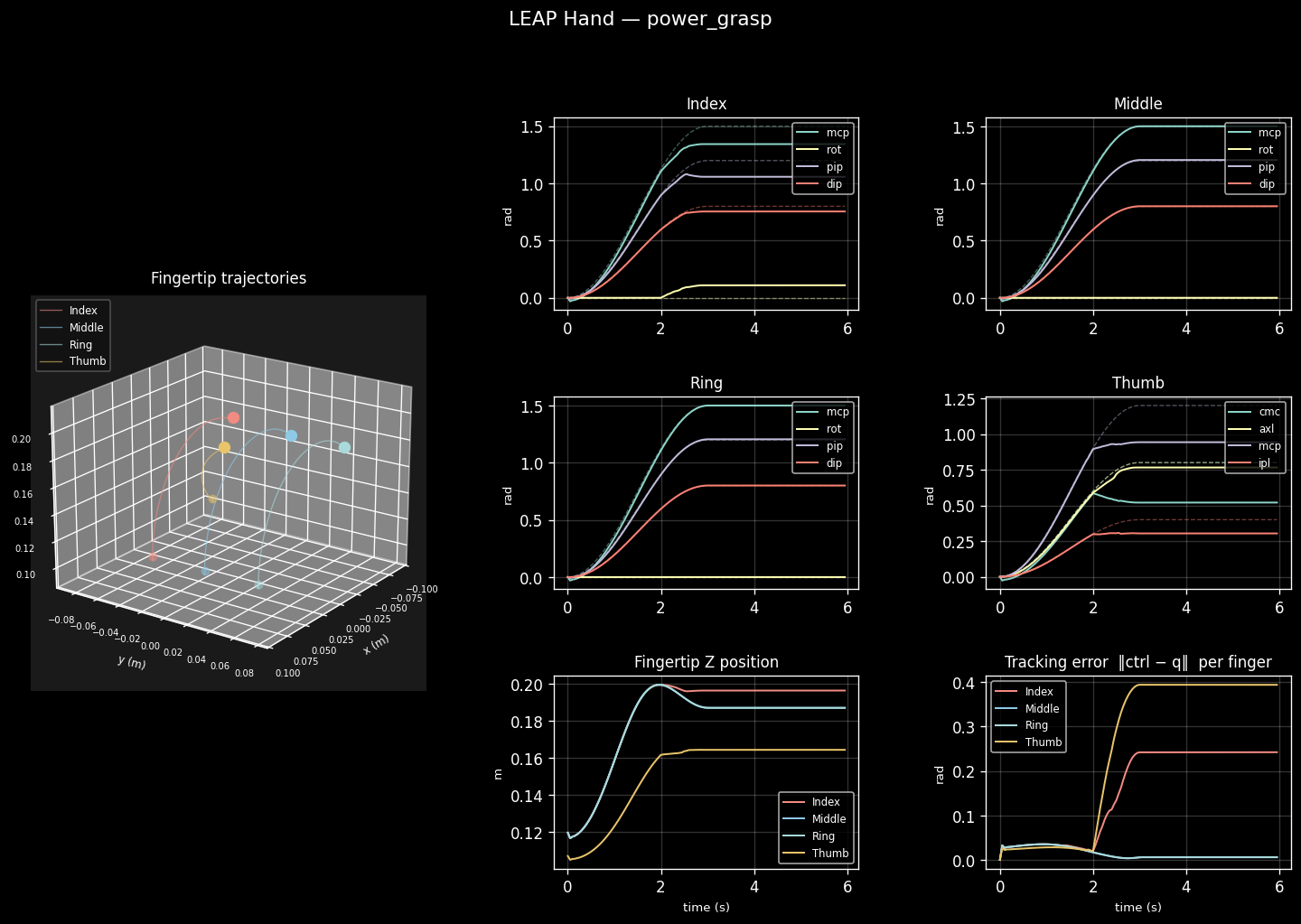} \\
{\scriptsize (a) Open/close} &
{\scriptsize (b) Precision pinch} &
{\scriptsize (c) Power grasp with disturbance}
\end{tabular}
\caption{Representative 16-DOF LEAP Hand MuJoCo simulation dashboards. The plots visualize joint references, measured joint 
trajectories, contact-equilibrium offsets, and object/grasp behaviour for three of the four evaluated grasp tasks; Hook is 
summarized in Table~\ref{tab:leapresults}.}
\label{fig:leaptasks}
\end{figure*}

Unlike the stiff single finger ($Q_{\text{pos}} = 10^8$), the 16-DOF hand runs soft impedance ($Q_{\text{pos}} = 2\times10^3$) with the conservative torque bounds of Table~\ref{tab:perjoint} for compliant grasping; per-joint tracking is intentionally loose, and during grasps the fingers settle to \emph{contact equilibria}---the commanded closed pose is partly unreachable through the grasped object, a physically correct offset rather than a tracking failure. Four grasps (Open/Close, Precision Pinch, Power, Hook) with cosine-interpolated trajectories are evaluated; Table~\ref{tab:leapresults} spans them (low end: gentle motions; high end: heavy-contact Power/Hook), and Fig.~\ref{fig:leaptasks} shows representative full-hand simulation traces. A 25\,mm, 50\,g sphere is held stably in all four configurations. These experiments test grasp formation, contact-equilibrium holding, and disturbance recovery; they do not claim solved in-hand object reorientation, which remains a separate manipulation benchmark.

\begin{table}[!t]
\caption{16-DOF LEAP Hand Results (Per-Joint, Across Four Grasps)}
\label{tab:leapresults}
\centering
\footnotesize
\setlength{\tabcolsep}{4pt}
\begin{tabular}{lccc}
\toprule
Joint group & \shortstack{Track RMSE\\(mrad)} & \shortstack{Settled offset\\(mrad)} & \shortstack{Hold std\\(mrad)} \\
\midrule
Finger MCP ($\times$3) & 6--465 & 0--871 & 0--338 \\
Finger ROT ($\times$3) & 0--13 & 0--29 & 0--11 \\
Finger PIP ($\times$3) & 2--263 & 0--238 & 0--172 \\
Finger DIP ($\times$3) & 0--133 & 0--341 & 0--88 \\
Thumb CMC / MCP / IPL & 9--164 & 4--209 & 0--142 \\
Thumb AXL & 12--79 & 1--80 & 0--72 \\
\bottomrule
\end{tabular}
\end{table}

The demonstrated multi-finger properties are stable grasping and disturbance rejection: a 2.5\,N downward force---a slipping-load 
surrogate---is applied during the Power and Hook hold phases, to the middle phalanges at Power (so as not to reseat the 
fingertip--thumb contact) and to the fingertips at Hook. The push deflects the hand by 205 / 506\,mrad (16-joint deviation norm, 
Power/Hook) and partially reseats the grasp: the contacting joints settle to a shifted contact equilibrium rather than returning 
to the pre-push pose. Measured against that post-release equilibrium, the transient decays below 15\,mrad within 0.49\,s (Power) 
and 0.71\,s (Hook) (Fig.~\ref{fig:leaptasks}; quantitative thumb metrics in Section~\ref{sec:leapkalman}). At the power-grasp 
pose the curled fingers contact the thumb; the resulting ${\sim}3$\,Nm constraint torques are not in the per-joint model, so 
thumb joints settle 60--210\,mrad from the geometric target at the contact equilibrium---physically correct---with bounded, 
non-divergent hold jitter that the contact-aware estimator of Section~\ref{sec:leapkalman} suppresses.

\emph{In-grasp knob rotation diagnostic.} To address whether the controller can do more than static grasp holding, we also 
evaluate a fixtured-knob manipulation task in the same MuJoCo model. The hand first power-grasps a 25\,mm-radius knob on a 
horizontal hinge, settles into contact, and then builds IK waypoints by rotating the settled fingertip contact points about 
the measured hinge axis. This settled-contact reference is important: a previous nominal pre-contact reference produced 
near-zero knob motion because the fingers jammed rather than rolled. Table~\ref{tab:knob} reports the corrected benchmark, 
using the coupled thumb block estimator to suppress thumb contact chatter. The 500\,Hz Impedance MPC+Kalman reaches 
38.6$^\circ$ of a commanded 40$^\circ$ rotation with 1.4$^\circ$ final error and 43.8\,N peak contact force, while 
admittance fails the task and produces large penetration forces. Crucially, it achieves the smallest final error and the 
lowest peak force among the passing controllers---roughly $4.4\times$ below stiff impedance (190.6\,N) and $1.8\times$ 
below PI (79.8\,N)---rolling the knob through gentler, safer contact rather than by gripping harder. This remains a simulation 
diagnostic, but it directly exercises in-grasp object motion rather than only pose holding.

\begin{table}[!t]
\caption{Fixtured-Knob In-Grasp Rotation Diagnostic}
\label{tab:knob}
\centering
\footnotesize
\setlength{\tabcolsep}{4pt}
\begin{tabular}{lccc}
\toprule
Controller & \shortstack{Achieved\\(deg)} & \shortstack{Hold error\\(deg)} & \shortstack{Peak force\\(N)} \\
\midrule
Stiff impedance & 36.2 & 3.8 & 190.6 \\
Admittance & -10.4 & 50.4 & 89792.1 \\
PI impedance & 33.5 & 6.5 & 79.8 \\
\textbf{Impedance MPC+Kalman} & \textbf{38.6} & \textbf{1.4} & \textbf{43.8} \\
\bottomrule
\end{tabular}
\end{table}

\section{Conclusion}
\label{sec:conclusion}

This paper argued that dexterous manipulation should be designed at the level of
\emph{interaction dynamics}, and developed
the constant-$A_d$ offset-free Impedance MPC architecture of \cite{cao2026phri}
into an actuator-agnostic simulation framework that makes those dynamics explicit
for dexterous fingers, with the hydraulic platform as the worked example: a
two-parameter feedforward reduces tendon transmissions to a residual double
integrator, and platform limits enter the QP as soft-relaxed constraints while
preserving the base architecture's nominal assumptions. The benchmarks established that (1) the
no-Kalman MPC realizes a rate-dependent first-move stiffness (18\,Nm/rad at
100\,Hz, 323\,Nm/rad at 500\,Hz, independently verified) that already beats
classical impedance; (2) the encoder-only Kalman estimator removes the residual
offset (0.1\,mrad SS, 153$\times$ RMS improvement); and (3) for this plant and
tuning, 500\,Hz is necessary and sufficient to keep both contact-onset and
release transients within the 15\,mrad precision window---the release transient
being a 100\,Hz failure mode invisible in cyclic metrics. The 16-DOF LEAP Hand
extension showed the per-joint architecture scales in MuJoCo without a full
rigid-body model, across four grasping tasks, two disturbance-recovery
experiments, and a fixtured-knob in-grasp rotation diagnostic.

Finally, the constant-$A_d$ backbone is one instance of a broader interaction-dynamics program that also underlies
fixed-base pHRI \cite{cao2026phri}, floating-base whole-body control \cite{caowbc}, and multi-agent motion planning
\cite{caodriving}. In a learning-augmented (\emph{physical-AI}) manipulation stack it is the model-based, certifiable
core: because contact is captured in closed form, a learned world model need only represent the residual
uncertainty---object properties, slip, and grasp intent---rather than re-learning finger dynamics, while the
hard-constrained QP supplies the force and actuation safety a learned policy cannot certify. We claim the
interaction-dynamics realization here, not a physical-AI system; that integration is future work.

\balance


\end{document}